\documentclass{article}

\PassOptionsToPackage{numbers, compress}{natbib}

\usepackage[preprint]{neurips_2021}

\usepackage[utf8]{inputenc} 
\usepackage[T1]{fontenc}    
\usepackage{hyperref}       
\usepackage{url}            
\usepackage{booktabs}       
\usepackage{amsfonts}       
\usepackage{amsmath}
\usepackage{amssymb}
\usepackage{nicefrac}       
\usepackage{microtype}      
\usepackage{adjustbox}
\usepackage{booktabs}
\usepackage{multirow}
\usepackage{adjustbox}
\usepackage{xcolor}
\usepackage{graphics,psfrag}
\usepackage{diagbox} 
\usepackage{subfigure}
\usepackage{bbding}
\usepackage{amssymb}
\usepackage{times}
\usepackage{epsfig}
\usepackage{wrapfig}
\usepackage{float}
\usepackage{rotating}
\usepackage{makecell}
\usepackage{pifont}

\definecolor{ForestGreen}{rgb}{0.13, 0.55, 0.13}
\definecolor{Maroon}{rgb}{0.69, 0.19, 0.0}

\newcolumntype{C}[1]{>{\centering\arraybackslash}m{#1}}
\newcolumntype{R}[1]{>{\raggedleft\arraybackslash}m{#1}}
\newcolumntype{P}[1]{>{\raggedright\arraybackslash}p{#1}}

\newcolumntype{M}[1]{>{\centering\arraybackslash}m{#1}}
\usepackage{makecell} 
\usepackage[usestackEOL]{stackengine}
\usepackage{graphicx}
\usepackage{capt-of}
\usepackage{booktabs}
\usepackage{varwidth}

\usepackage{caption}
\usepackage{subcaption}
\usepackage{xspace}

\usepackage{import}
\usepackage{animate}
\usepackage{arydshln}
\usepackage{multirow}
\usepackage[normalem]{ulem}

\usepackage{algpseudocode}

\title{AMD-Hummingbird: Towards an Efficient Text-to-Video Model}

\def\authorBlock{
Takashi Isobe$^1$ \qquad
He Cui$^1$ \qquad
Dong Zhou$^{1}$ \\
Mengmeng Ge$^{1,2}$ \qquad
Dong Li$^{1}$ \qquad
Emad Barsoum$^{1}$ \\
$^1$ Advanced Micro Devices, Inc.\\
$^2$ Tsinghua University \\
Homepage: \url{https://www.amd.com/en/developer/resources/technical-articles.html} \\
Github: \url{https://github.com/AMD-AIG-AIMA/AMD-Hummingbird-T2V} 
}

\author{\authorBlock}

\makeatletter

\begin{document}

\maketitle
\renewcommand{\thefootnote}{\fnsymbol{footnote}} 

\begin{abstract}
Text-to-Video (T2V) generation has attracted significant attention for its ability to synthesize realistic videos from textual descriptions. However, existing models struggle to balance computational efficiency and high visual quality, particularly on resource-limited devices, \textit{e.g.,} iGPUs and mobile phones. Most prior work prioritizes visual fidelity while overlooking the need for smaller, more efficient models suitable for real-world deployment. To address this challenge, we propose a lightweight T2V framework, termed Hummingbird, which prunes existing models and enhances visual quality through visual feedback learning. Our approach reduces the size of the U-Net from 1.4 billion to 0.7 billion parameters, significantly improving efficiency while preserving high-quality video generation. Additionally, we introduce a novel data processing pipeline that leverages Large Language Models (LLMs) and Video Quality Assessment (VQA) models to enhance the quality of both text prompts and video data. To support user-driven training and style customization, we publicly release the full training code, including data processing and model training. Extensive experiments show that our method achieves a 31× speedup compared to state-of-the-art models such as VideoCrafter2, while also attaining the highest overall score on VBench. Moreover, our method supports the generation of videos with up to 26 frames, addressing the limitations of existing U-Net-based methods in long video generation. Notably, the entire training process requires only four GPUs, yet delivers performance competitive with existing leading methods. Hummingbird presents a practical and efficient solution for T2V generation, combining high performance, scalability, and flexibility for real-world applications.

\end{abstract}

\section{Introduction}
Text-to-video (T2V) generation~\cite{chen2023videocrafter1,chen2024videocrafter2,guo2023animatediff,li2024t2v,lin2024open,pikalabs2023,open-sora,zhang2023show1,ding2022cogview2,hong2022cogvideo,khachatryan2023text2video,blattmann2023align,ho2022imagen,singer2022make,wang2023modelscope} has attracted significant attention for its ability to
synthesize realistic videos from textual descriptions. While most existing methods focuses on improving visual quality, the efficiency of T2V diffusion models remains underexplored.In practical deployment scenarios, especially on resource-constrained devices such as iGPUs and NPUs, model size and inference speed are critical factors. Achieving both computational efficiency and high visual performance remains a significant challenge. Recent efforts~\cite{wang2023videolcm,wang2024animatelcm,li2024t2v,li2024t2v2} aim to improve efficiency by reducing the number of inference steps. For instance, VideoLCM~\cite{wang2023videolcm} introduces a self-consistency loss in the latent space to align model performance across different step counts. AnimateLCM~\cite{wang2024animatelcm} uses consistency distillation to accelerate generation. Despite these advances, the models remain large, which makes deployment on edge devices difficult. One na\"ive solution is to prune the network structure. However, this often leads to performance degradation because it disrupts the learned diffusion priors, even when followed by fine-tuning. 

To address this challenge, we propose a two-stage diffusion model distillation pipeline that first prunes model parameters and then restores visual quality through visual feedback learning. Specifically, our method reduces the U-Net size from 1.4 billion parameters, as seen in the widely used VideoCrafter2~\cite{chen2024videocrafter2}, to 0.7 billion. This enables high-quality video generation with minimal inference steps. The resulting efficient T2V model, named \textit{AMD-Hummingbird}, achieves a 31$\times$ speedup over VideoCrafter2 when deployed on AMD Instinct\texttrademark{} MI250 accelerators. In addition, on a consumer laptop equipped with an iGPU (Radeon\texttrademark{} 880M) and CPU (Ryzen\texttrademark{} AI 9 365), the model generates a 26-frame video in just 50 seconds\footnote{Tested in Nov 2024 by AMD using AMD Radeon\texttrademark{} 880M and Ryzen\texttrademark{} AI 9 365 on Ubuntu 6.8.0-51-generic.}. We also introduce a novel data processing pipeline that leverages large language models (LLMs) to recaption text prompts and select high-quality video samples, thereby optimizing training and improving the model performance. The proposed Hummingbird highlights AMD's significant advancements in AI-driven video generation. The main contribution can be summarized as follows: 
\begin{itemize}

\item We propose an innovative network pruning strategy for T2V models that significantly reduces model parameters while maintaining visual quality through visual feedback learning. The proposed method can generate high-quality long videos with up to 26 frames.

\item We introduce a novel data processing pipeline that leverages large language models (LLMs) to recaption text prompts and employs video quality assessment (VQA) models to select high-quality video samples for training.

\item We publicly release the training code and data processing pipeline, enabling users to easily customize the style of generated videos. This provides flexibility to adapt models to specific needs and preferences.

\item Extensive experiments demonstrate that our method achieves a 31$\times$ speedup over state-of-the-art models such as VideoCrafter2, while also achieving the highest overall score on VBench.

\end{itemize}

\section{Motivation}

The development of Hummingbird is driven by several key factors. Firstly, efficiency plays a crucial role. Existing open-source T2V models like VideoCrafter2~\cite{chen2024videocrafter2} and CogVideo~\cite{hong2022cogvideo} have made significant strides in visual performance, but they do not yet achieve efficient generation on typical GPUs. There is a critical need for smaller models that align with operational constraints while maintaining high performance. 

Another important consideration is model size and inference speed. Most current research emphasizes improving visual performance, often at the expense of model size and inference speed, which are crucial for practical deployment. AMD approach focuses on balancing these aspects to ensure both high-quality outputs and efficient operations. Customization is also a key advantage of building our own model. The proposed structural distillation method allows for adaptable and reward-optimized models, providing users with tailored solutions that meet their specific needs. 

Data quality is essential for effective model learning. This novel data processing pipeline helps ensure that the model is trained on top-quality visual and textual data, enhancing its performance and reliability. High-quality training data is critical for the model to generate richer textural details and better understand textual input. Finally, by open sourcing the training code, dataset, and model weights, AMD supports the AI community. This enables developers to replicate and explore T2V generation on AMD AI platforms, fostering an open and collaborative approach to AI development, and ensuring that the benefits of AI advancements are widely shared.

\section{Method}

\begin{figure*}[h]
	\centering
	\includegraphics[width=0.9\textwidth]{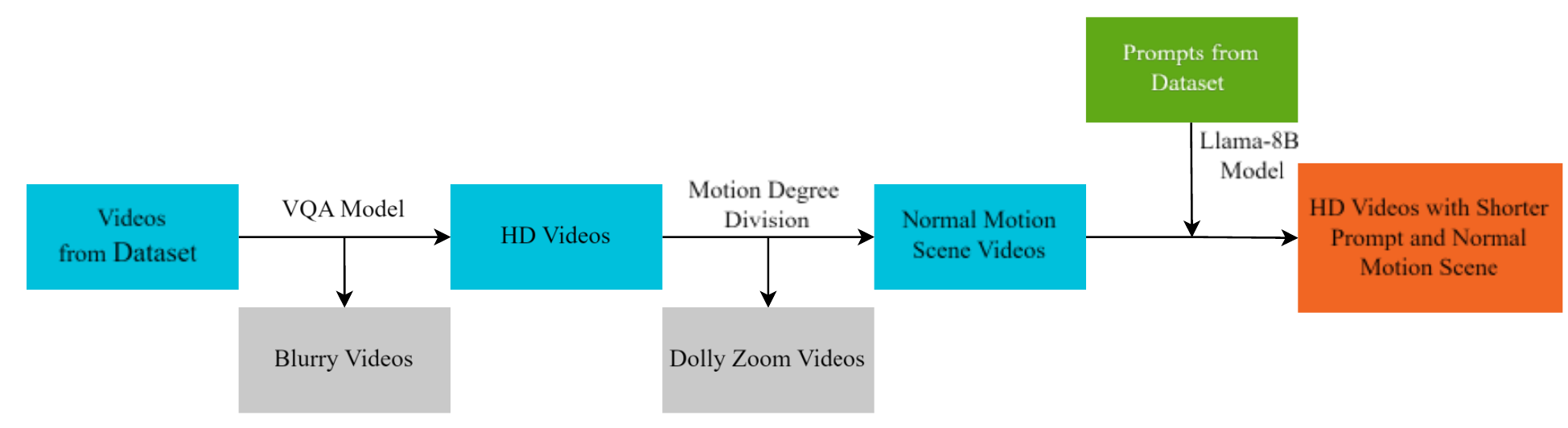}
	\caption{ Illustration of the proposed data processing pipeline, which includes video quality assessment, motion filtering, and prompt re-captioning using large language models to improve training data quality.}
	\label{fig:pred_mask}
\end{figure*}

Building on these considerations, we propose an efficient T2V model along with a novel two-stage structural pruning pipeline that first reduces network parameters and then enhances visual quality through feedback learning. In addition, we introduce a data processing pipeline designed to improve the quality of both text prompts and video samples.

\subsection{Data Processing Pipeline}

The proposed data processing pipeline includes video quality assessment, motion estimation, and prompt re-captioning. Low-quality video data can slow the convergence of the loss function during training and reduce the performance of the T2V diffusion model. To address this issue, the pipeline begins by applying Video Quality Assessment (VQA) models~\cite{wu2023exploring} to evaluate each video's quality based on aesthetic and compression-related metrics, as shown in Figure~\ref{fig:pred_mask}. A motion estimation module is then used to filter out dolly zoom videos from the set of high-quality samples. To improve the quality of textual inputs, the pipeline leverages the prior knowledge of a large language model, specifically LLaMA-8B~\cite{grattafiori2024llama}, to perform prompt re-captioning. Experimental results show that training on videos with high visual and textual quality enables the T2V diffusion model to generate more detailed textures and better interpret textual input.

\begin{figure*}[t]
	\centering
	\includegraphics[width=0.9\textwidth]{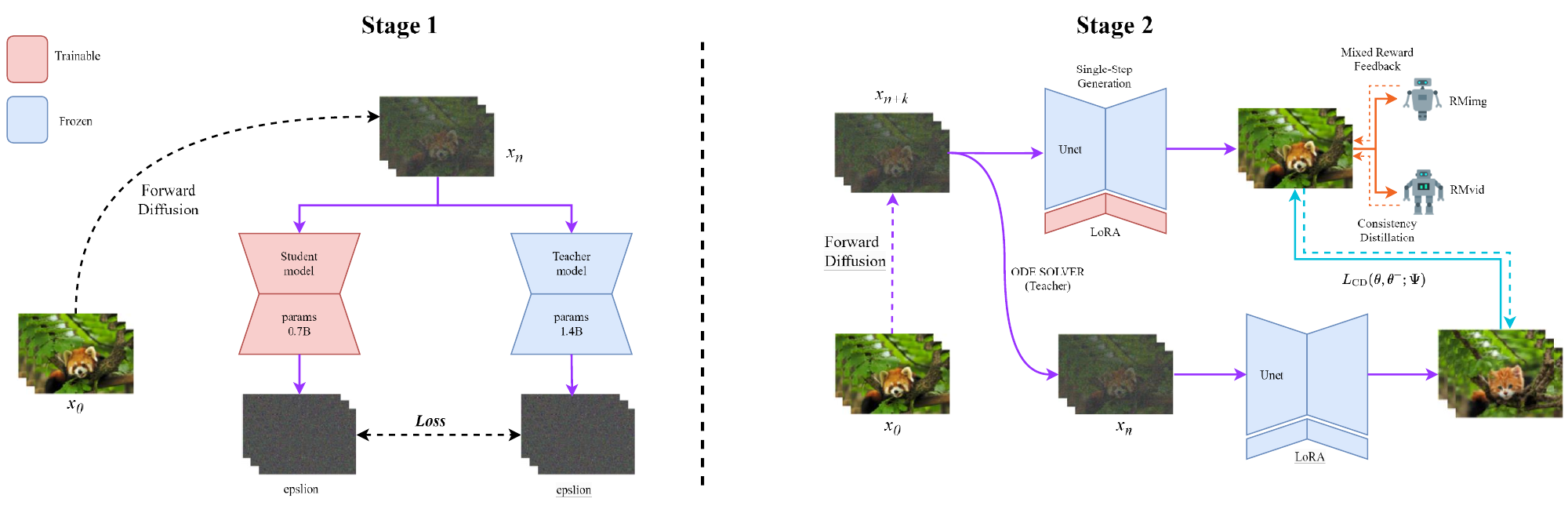}
	\caption{Illustration of the proposed two-stage T2V diffusion model distillation pipeline. The first stage prunes the model's parameters to improve efficiency, while the second stage enhances visual quality through feedback learning.}
	\label{fig:pred_mask}
\end{figure*}

\subsection{Network Pruning}
Model distillation~\cite{luhman2021knowledge,salimans2022progressive,meng2023distillation} is a proven approach for reducing model size, making it suitable for deployment on resource-constrained hardware. However, it has not been widely explored in the context of text-to-video (T2V) diffusion models. The main challenge lies in effectively transferring knowledge from a large, high-performing model to a smaller one without causing significant performance degradation. To overcome this, we introduce a two-stage distillation framework designed specifically for T2V diffusion models.

\textbf{First Stage:} The first step involves constructing a smaller model by reducing the number of blocks by half in each layer and removing all middle blocks from the U-Net architecture. Maintaining structural similarity with the original model during this pruning process is essential to support better adaptation of diffusion priors and significantly accelerates training. We also explored various alternative architectural configurations and found that several pruned variants achieved comparable performance after fine-tuning. Based on these findings, we do not apply any task-specific architectural design at this stage. Taking VideoCrafter2~\cite{chen2024videocrafter2} as an example, which contains approximately 1.4 billion parameters, our pruned version reduces the parameter count to 0.7 billion. We then fine-tune the smaller model using the outputs of the original model as supervision, following by~\cite{luo2023latent,wang2023videolcm,song2023consistency,song2023improved}. The training details are listed below:
\begin{itemize}
\item  The model is trained directly on the video outputs generated by the original model after classifier-free guidance (CFG)~\cite{ho2022classifier} is applied.
\item  We minimize the difference between adjacent points along the Probability Flow Ordinary Differential Equation (PF ODE) trajectory using numerical solvers.
\end{itemize}

\textbf{Second Stage:} The second stage focuses on further enhancing the visual quality of the Hummingbird diffusion model, with improvements in tonal range, color saturation, textural detail, and overall visual fidelity. A straightforward approach would be to collect or synthesize additional video data that features high-quality visual effects. For example, some methods utilize video clips from professionally produced films, which benefit from high-end cinematography and extensive post-production. However, acquiring such data is time-consuming, resource-intensive, and introduces practical limitations related to bandwidth usage and data availability.

To address these challenges, we propose reusing the training data from the first stage to further improve visual quality. Inspired by~\cite{li2024t2v2}, we leverage reward feedback from a mixture of reward models to extract valuable visual cues from the existing dataset. This feedback-guided approach enables the model to refine fine-grained visual features without requiring new data. To enhance temporal coherence and visual consistency, we incorporate both image-text and video-text reward models during training. The second-stage training is highly efficient and requires only one GPU-day, utilizing four AMD Instinct MI250 GPUs with 64 GB of memory each.

\section{Experiment}
\subsection{Experimental Setting}
\textbf{Training Details}

\begin{table}[h]
\centering
\setlength\tabcolsep{3pt}
\begin{center}
\caption{\textbf{VBench~\citep{huang2023vbench} Evaluation Results by Dimension.} This table compares the performance of 11 video generation models across the 16 individual dimensions defined in VBench. A higher score indicates better performance for each dimension. The best result for each dimension is shown in \textbf{bold}, and the second-best result is \underline{underlined}.}
\resizebox{\linewidth}{!}{
\begin{tabular}{l|c||c|c|c|c|c|c|c|c}
\toprule
\textbf{Models}  & \textbf{\Centerstack{Total\\Score}} & \textbf{\Centerstack{Quality\\Score}} & {\Centerstack{Subject\\Consist.}} & {\Centerstack{BG \\Consist.}} & 
{\Centerstack{Temporal\\Flicker.}} & {\Centerstack{Motion\\Smooth.}} & {\Centerstack{Aesthetic\\Quality}} & {\Centerstack{Dynamic\\Degree}} & {\Centerstack{Image\\Quality}} \\
\midrule
ModelScopeT2V & 75.75 & 78.05 & 89.87 & 95.29 & 98.28 & 95.79 & 52.06 & \underline{66.39}  &58.57 \\ 
LaVie & 77.08 & 78.78 & 91.41 & 97.47 & 98.30 & 96.38 & 54.94 & 49.72  &61.90 \\ 
Show-1 & 78.93 & 80.42 & 95.53 & 98.02 & 99.12 & 98.24 & 57.35 & 44.44  &58.66 \\ 
VideoCrafter1 & 79.72 & 81.59 & 95.10 & 98.04 & 98.93 & 95.67 & 62.67 & 55.00  &65.46 \\ 
Pika & 80.40 & 82.68 & 96.76 & \textbf{98.95} & \textbf{99.77} & \underline{99.51} & 63.15& 37.22  &62.33 \\ 
VideoCrafter2 & 80.44 & 82.20 & 96.85 & \underline{98.22} & 98.41 & 97.73 & 63.13 & 42.50  &67.22 \\ 
Gen-2 & 80.58 & 82.47 & \textbf{97.61} & 97.61 & \underline{99.56} & \textbf{99.58} & 66.96 & 18.89  &67.42 \\ 
AnimateLCM &77.74  &80.68  &96.57  &96.57  &98.41  &98.33  &63.26  &33.33  &62.30 \\ 
VideoLCM &73.27  &77.65 &96.55  &97.23  &97.33  &97.01  &59.93  &5.56  &66.43 \\ 
T2V-Turbo  & \underline{81.01} & 82.57 & 96.28 & 97.02 & 97.48 & 97.34 & 63.04 & 49.17  &\textbf{72.49}  \\ 
Hummingbird 16 frames (Ours) &\textbf{81.35}   &\textbf{83.73} &95.87  &96.77  &95.24 &96.14  &\textbf{68.04}  &\textbf{79.17}  &\underline{71.04}  \\ 
Hummingbird 26 frames (Ours) &80.31   &\underline{83.11} &\underline{96.97}  &97.73  &97.64 &96.97  &\underline{67.82}  &50.00  &69.94  \\ 
\end{tabular}
}
\vspace{3ex}
\resizebox{\linewidth}{!}{
\begin{tabular}{l|c|c|c|c|c|c|c|c|c|c}
\toprule
\textbf{Models}  &  \textbf{\Centerstack{Semantic\\Score}}  &{\Centerstack{Object\\Class}}  & {\Centerstack{Multiple\\Objects}} & {\Centerstack{Human\\Action}} & {Color} & {\Centerstack{Spatial\\Relation.}} & {Scene} & {\Centerstack{Appear.\\Style}} & {\Centerstack{Temporal\\Style}} & {\Centerstack{Overall\\Consist.}} \\
\midrule
ModelScopeT2V &  66.54 &82.25 & 38.98 & 92.40 & 81.72 & 33.68 & 39.26 & 23.39 & 25.37 & 25.67 \\ 
LaVie &  70.31 &91.82 & 33.32 & \textbf{96.80} & 86.39 & 34.09 & 52.69 & 23.56 & \textbf{25.93} & 26.41 \\ 
Show-1 &  72.98 &93.07 & 45.47 & \underline{95.60} & 86.35 & 53.50 & 47.03 & 23.06 & 25.28 & 27.46 \\ 
VideoCrafter1 &  72.22 &78.18 & 45.66 & 91.60 & \underline{93.32} & 58.86 & 43.75 & 24.41 & 25.54 & 26.76 \\ 
Pika &  71.26 &87.45 & 46.69 & 88.00 & 85.31 & \underline{65.65} & 44.80 & 21.89 & 24.44 & 25.47 \\ 
VideoCrafter2 &  \underline{73.42} &92.55 & 40.66 & 95.00 &92.92& 35.86 & \underline{55.29} & \textbf{25.13} & \underline{25.84} & \textbf{28.23} \\ 
Gen-2 &  73.03 &90.92 & \textbf{55.47} & 89.20 & 89.49 & \textbf{66.91} & 48.91 & 19.34 & 24.12 & 26.17 \\ 
AnimateLCM &66.00  &87.34  &34.68  &83.00  &85.62  &40.86  &46.29  &19.87  &24.19 &25.57\\ 
VideoLCM &55.75  &75.40  &12.50  &73.00  &82.64  &19.85  &35.10  &19.87  &22.25 &23.68  \\ 
T2V-Turbo &  \textbf{74.76} &93.96 & \underline{54.65} & 95.20 & 89.90 & 38.67 & \textbf{55.58} &\underline{24.42} & 25.51 & \underline{28.16}\\ 
Hummingbird 16 frames(Ours) &71.84 &\textbf{96.36} &35.44 &94.00  &91.63  &38.96 &52.91  &22.43  &25.53  &28.09 \\ 
Hummingbird 26 frames(Ours) &69.10 &\underline{94.86} &33.99 &92.00  &\textbf{94.79}  &26.87 &49.49  &21.90  &24.24  &27.65 \\ 
\bottomrule
\end{tabular}
}

\vspace{-20pt}
\label{tab:auto-eval}
\end{center}
\end{table}

\begin{table*}[ht]
\centering
\caption{\textbf{Quantitative Comparison of Different Models.} Latency is measured on the AMD Instinct MI250 accelerator using the FP16 data type.}
\setlength{\tabcolsep}{4pt}
\resizebox{1\textwidth}{!}{
\begin{tabular}{c c c c c c c}
\hline
\textbf{Metric} & \textbf{VideoLCM} & \textbf{AnimateLCM} & \textbf{T2V-Turbo}  & \textbf{VideoCrafter2} & \textbf{Hummingbird (Ours)} \\
\hline
Steps             & 4  & 4  & 4   & 50  & 4  \\
Unet Params (B)       & 1.4  & \underline{1.3}   & 1.4  & 1.4 & \textbf{0.7}  \\
Inference Latency (s) &\underline{2.4}  & 6.4 & 2.5   & 44.2  & \textbf{1.3}  \\
\hline
\end{tabular}
}
\label{tab:comparison_new}
\end{table*}

We train Hummingbird on the WebVid-10M~\cite{webvid} dataset. WebVid is a publicly available, large-scale dataset consisting of 10 million video-text pairs collected from stock footage platforms. It features a wide range of motion patterns and diverse scenes, making it suitable for general-purpose T2V training. A subset of high-quality video data is selected using our proposed data processing pipeline for the first-stage training. For the second-stage training, a smaller subset is randomly sampled from this selection. The first stage is trained for 200K steps with a learning rate of $1\times10^{-4}$ and a batch size of 16. The second stage is trained for 80K steps with a learning rate of $1\times10^{-5}$ and a batch size of 8. During training, we randomly sample 16 frames per video at a resolution of $320\times512$, using a dynamic frame stride. The frame rate condition varies depending on the codec of the original video. Training is conducted on AMD Instinct MI250 GPUs, which are based on the CDNA~2 architecture and feature 362.1 TFLOPs of peak FP16 performance, 64~GB of HBM2e memory, 3.2~TB/s peak memory bandwidth, and a 500W TDP. The software stack includes Python~3.8, ROCm\texttrademark~5.6.0, PyTorch~2.2.0, and FlashAttention~2.2.0. It is worth noting that we also possess a privately collected dataset that can further enhance model performance. However, due to licensing and privacy constraints, this dataset cannot be released. Therefore, all training for the public code release is conducted using the publicly available WebVid-10M dataset.

\textbf{Evaluation Metrics}
We evaluate the proposed method on the standard benchmark VBench~\cite{} which consists of 16 disentangled dimensions and is designed to comprehensively evaluate T2V models. Followed by~\cite{li2024t2v2}, we also present the Quality Score and Semantic Score. The Quality Score is computed as a weighted sum of seven normalized sub-metrics: Subject Consistency~\cite{caron2021emerging}, Background Consistency~\cite{radford2021learning}, Temporal Flickering, Motion Smoothness~\cite{li2023amt}, Aesthetic Quality~\cite{schuhmann2022laion}, Dynamic Degree~\cite{teed2020raft}, and Image Quality~\cite{ke2021musiq}. The Semantic Score is obtained by averaging nine normalized sub-metrics: Object Class~\cite{wu2022grit}, Multiple Object~\cite{wu2022grit}, Human Action~\cite{li2023unmasked}, Color~\cite{wu2022grit}, Spatial Relationship~\cite{huang2023t2i}, Scene~\cite{huang2023tag2text}, Appearance Style~\cite{wang2023internvid}, Temporal Style~\cite{wang2023internvid}, and Overall Consistency~\cite{wang2023internvid}. The Total Score is then calculated as a weighted sum of the Quality Score and Semantic Score, providing an overall measure of the model’s visual and semantic performance.

\begin{figure}[h]
    \centering
    \animategraphics[width=\linewidth]{10}{video/}{0001}{0020}
     \caption{\textbf{Qualitative Results.} Video results generated from various text prompts. The proposed Hummingbird model is capable of generating high-quality visual results while accurately following the text prompts and producing semantically consistent content. We recommend viewing the video results using Adobe Acrobat Reader.}
     \label{fig:video_demos}
    \vspace{-10pt}
    
\end{figure}

\subsection{Comparison with State-of-the-arts}
\textbf{Quantitative comparison.}
To further evaluate the visual performance of Hummingbird, we compare it against 10 state-of-the-art models, including both open-source and closed-source foundation models. The evaluation is conducted using text prompts from VBench~\cite{huang2023vbench}, which cover a wide range of scenes such as natural landscapes, science fiction environments, and urban settings. The models considered in the comparison include ModelScope~\cite{wang2023modelscope}, LaVie~\cite{wang2023lavie}, Show-1~\cite{zhang2023show1}, VideoCrafter1~\cite{chen2023videocrafter1}, Pika~\cite{pikalabs2023}, VideoCrafter2~\cite{chen2024videocrafter2}, Gen-2~\cite{gen2}, AnimateLCM~\cite{wang2024animatelcm}, VideoLCM~\cite{wang2023videolcm}, and T2V-Turbo~\cite{li2024t2v}. As shown in Table~\ref{tab:auto-eval}, Hummingbird, despite its smaller size, achieves the highest total score among all models, demonstrating its effectiveness in generating high-quality video content with strong computational efficiency. In addition, the proposed Hummingbird model is capable of generating long video sequences with up to 26 frames while maintaining the same performance.

\textbf{Inference speed.}
Table~\ref{tab:comparison_new} highlights the inference efficiency of Hummingbird. When evaluated on the AMD Instinct MI250 accelerator, our model achieves approximately a 31$\times$ speedup compared to VideoCrafter2. This result illustrates Hummingbird’s strong balance of speed, computational efficiency, and visual fidelity, setting a new baseline for lightweight text-to-video generation models. 

\textbf{Qualitative results.} Figure~\ref{fig:video_demos} presents qualitative examples generated by Hummingbird. The model is capable of producing visually appealing videos with strong temporal consistency. Furthermore, it demonstrates improved alignment between the generated video content and the semantic details of the input text prompts. For optimal viewing quality, we recommend using Adobe Acrobat.

\section{Future Plan}
We are preparing to release a lightweight image-to-video (I2V) model that supports long video generation with up to 26 frames. This model is designed to offer efficient and high-quality video synthesis from static images, making it suitable for real-world applications on resource-constrained devices. In parallel, we are actively developing lightweight DiT-based models for both T2V and I2V generation. While these models currently rely on privately collected datasets for training, we aim to adapt them to work with publicly available datasets. This effort will improve reproducibility and accessibility for the broader research community.

\section{Conclusion}
The AMD Hummingbird T2V diffusion model marks a significant advancement in text-to-video generation, leveraging the capabilities of AMD state-of-the-art hardware and software technologies. The structure distillation method proposed for the T2V diffusion model exemplifies the dedication of AMD to fostering AI innovation, offering exceptional performance and efficiency. Additionally, AMD hardware and software excel in training and inferencing large-scale models like T2V diffusion model, making it a key player in advancing AI and empowering developers to push the boundaries of innovation. 
{\small
\bibliographystyle{unsrt}
\bibliography{res}

\begin{thebibliography}{10}

\bibitem{chen2023videocrafter1}
Haoxin Chen, Menghan Xia, Yingqing He, Yong Zhang, Xiaodong Cun, Shaoshu Yang, Jinbo Xing, Yaofang Liu, Qifeng Chen, Xintao Wang, et~al.
\newblock Videocrafter1: Open diffusion models for high-quality video generation.
\newblock {\em arXiv preprint arXiv:2310.19512}, 2023.

\bibitem{chen2024videocrafter2}
Haoxin Chen, Yong Zhang, Xiaodong Cun, Menghan Xia, Xintao Wang, Chao Weng, and Ying Shan.
\newblock Videocrafter2: Overcoming data limitations for high-quality video diffusion models.
\newblock {\em arXiv preprint arXiv:2401.09047}, 2024.

\bibitem{guo2023animatediff}
Yuwei Guo, Ceyuan Yang, Anyi Rao, Zhengyang Liang, Yaohui Wang, Yu~Qiao, Maneesh Agrawala, Dahua Lin, and Bo~Dai.
\newblock Animatediff: Animate your personalized text-to-image diffusion models without specific tuning.
\newblock {\em arXiv preprint arXiv:2307.04725}, 2023.

\bibitem{li2024t2v}
Jiachen Li, Qian Long, Jian Zheng, Xiaofeng Gao, Robinson Piramuthu, Wenhu Chen, and William~Yang Wang.
\newblock T2v-turbo-v2: Enhancing video generation model post-training through data, reward, and conditional guidance design.
\newblock {\em arXiv preprint arXiv:2410.05677}, 2024.

\bibitem{lin2024open}
Bin Lin, Yunyang Ge, Xinhua Cheng, Zongjian Li, Bin Zhu, Shaodong Wang, Xianyi He, Yang Ye, Shenghai Yuan, Liuhan Chen, et~al.
\newblock Open-sora plan: Open-source large video generation model.
\newblock {\em arXiv preprint arXiv:2412.00131}, 2024.

\bibitem{pikalabs2023}
{Pika Labs}.
\newblock Accessed september 25, 2023, 2023.

\bibitem{open-sora}
Open-Sora.
\newblock Open-sora: Democratizing efficient video production for all, 2024.

\bibitem{zhang2023show1}
David~Junhao Zhang, Jay~Zhangjie Wu, Jia-Wei Liu, Rui Zhao, Lingmin Ran, Yuchao Gu, Difei Gao, and Mike~Zheng Shou.
\newblock Show-1: Marrying pixel and latent diffusion models for text-to-video generation.
\newblock {\em arXiv preprint arXiv:2309.15818}, 2023.

\bibitem{ding2022cogview2}
Ming Ding, Wendi Zheng, Wenyi Hong, and Jie Tang.
\newblock Cogview2: Faster and better text-to-image generation via hierarchical transformers.
\newblock 2022.

\bibitem{hong2022cogvideo}
Wenyi Hong, Ming Ding, Wendi Zheng, Xinghan Liu, and Jie Tang.
\newblock Cogvideo: Large-scale pretraining for text-to-video generation via transformers.
\newblock In {\em The Eleventh International Conference on Learning Representations}, 2022.

\bibitem{khachatryan2023text2video}
Levon Khachatryan, Andranik Movsisyan, Vahram Tadevosyan, Roberto Henschel, Zhangyang Wang, Shant Navasardyan, and Humphrey Shi.
\newblock Text2video-zero: Text-to-image diffusion models are zero-shot video generators.
\newblock In {\em Proceedings of the IEEE/CVF International Conference on Computer Vision}, pages 15954--15964, 2023.

\bibitem{blattmann2023align}
Andreas Blattmann, Robin Rombach, Huan Ling, Tim Dockhorn, Seung~Wook Kim, Sanja Fidler, and Karsten Kreis.
\newblock Align your latents: High-resolution video synthesis with latent diffusion models.
\newblock In {\em Proceedings of the IEEE/CVF Conference on Computer Vision and Pattern Recognition}, pages 22563--22575, 2023.

\bibitem{ho2022imagen}
Jonathan Ho, William Chan, Chitwan Saharia, Jay Whang, Ruiqi Gao, Alexey Gritsenko, Diederik~P Kingma, Ben Poole, Mohammad Norouzi, David~J Fleet, et~al.
\newblock Imagen video: High definition video generation with diffusion models.
\newblock {\em arXiv preprint arXiv:2210.02303}, 2022.

\bibitem{singer2022make}
Uriel Singer, Adam Polyak, Thomas Hayes, Xi~Yin, Jie An, Songyang Zhang, Qiyuan Hu, Harry Yang, Oron Ashual, Oran Gafni, et~al.
\newblock Make-a-video: Text-to-video generation without text-video data.
\newblock {\em arXiv preprint arXiv:2209.14792}, 2022.

\bibitem{wang2023modelscope}
Jiuniu Wang, Hangjie Yuan, Dayou Chen, Yingya Zhang, Xiang Wang, and Shiwei Zhang.
\newblock Modelscope text-to-video technical report.
\newblock {\em arXiv preprint arXiv:2308.06571}, 2023.

\bibitem{wang2023videolcm}
Xiang Wang, Shiwei Zhang, Han Zhang, Yu~Liu, Yingya Zhang, Changxin Gao, and Nong Sang.
\newblock Videolcm: Video latent consistency model.
\newblock {\em arXiv preprint arXiv:2312.09109}, 2023.

\bibitem{wang2024animatelcm}
Fu-Yun Wang, Zhaoyang Huang, Weikang Bian, Xiaoyu Shi, Keqiang Sun, Guanglu Song, Yu~Liu, and Hongsheng Li.
\newblock Animatelcm: Computation-efficient personalized style video generation without personalized video data.
\newblock In {\em SIGGRAPH Asia 2024 Technical Communications}, pages 1--5. 2024.

\bibitem{li2024t2v2}
Jiachen Li, Weixi Feng, Tsu-Jui Fu, Xinyi Wang, Sugato Basu, Wenhu Chen, and William~Yang Wang.
\newblock T2v-turbo: Breaking the quality bottleneck of video consistency model with mixed reward feedback.
\newblock {\em arXiv preprint arXiv:2405.18750}, 2024.

\bibitem{wu2023exploring}
Haoning Wu, Erli Zhang, Liang Liao, Chaofeng Chen, Jingwen Hou, Annan Wang, Wenxiu Sun, Qiong Yan, and Weisi Lin.
\newblock Exploring video quality assessment on user generated contents from aesthetic and technical perspectives.
\newblock In {\em Proceedings of the IEEE/CVF International Conference on Computer Vision}, pages 20144--20154, 2023.

\bibitem{grattafiori2024llama}
Aaron Grattafiori, Abhimanyu Dubey, Abhinav Jauhri, Abhinav Pandey, Abhishek Kadian, Ahmad Al-Dahle, Aiesha Letman, Akhil Mathur, Alan Schelten, Alex Vaughan, et~al.
\newblock The llama 3 herd of models.
\newblock {\em arXiv preprint arXiv:2407.21783}, 2024.

\bibitem{luhman2021knowledge}
Eric Luhman and Troy Luhman.
\newblock Knowledge distillation in iterative generative models for improved sampling speed.
\newblock {\em arXiv preprint arXiv:2101.02388}, 2021.

\bibitem{salimans2022progressive}
Tim Salimans and Jonathan Ho.
\newblock Progressive distillation for fast sampling of diffusion models.
\newblock In {\em International Conference on Learning Representations}, 2021.

\bibitem{meng2023distillation}
Chenlin Meng, Robin Rombach, Ruiqi Gao, Diederik Kingma, Stefano Ermon, Jonathan Ho, and Tim Salimans.
\newblock On distillation of guided diffusion models.
\newblock In {\em Proceedings of the IEEE/CVF Conference on Computer Vision and Pattern Recognition}, pages 14297--14306, 2023.

\bibitem{luo2023latent}
Simian Luo, Yiqin Tan, Longbo Huang, Jian Li, and Hang Zhao.
\newblock Latent consistency models: Synthesizing high-resolution images with few-step inference.
\newblock {\em arXiv preprint arXiv:2310.04378}, 2023.

\bibitem{song2023consistency}
Yang Song, Prafulla Dhariwal, Mark Chen, and Ilya Sutskever.
\newblock Consistency models.
\newblock {\em International conference on machine learning}, 2023.

\bibitem{song2023improved}
Yang Song and Prafulla Dhariwal.
\newblock Improved techniques for training consistency models.
\newblock In {\em The Twelfth International Conference on Learning Representations}, 2023.

\bibitem{ho2022classifier}
Jonathan Ho and Tim Salimans.
\newblock Classifier-free diffusion guidance.
\newblock In {\em NeurIPS 2021 Workshop on Deep Generative Models and Downstream Applications}, 2021.

\bibitem{huang2023vbench}
Ziqi Huang, Yinan He, Jiashuo Yu, Fan Zhang, Chenyang Si, Yuming Jiang, Yuanhan Zhang, Tianxing Wu, Qingyang Jin, Nattapol Chanpaisit, Yaohui Wang, Xinyuan Chen, Limin Wang, Dahua Lin, Yu~Qiao, and Ziwei Liu.
\newblock {VBench}: Comprehensive benchmark suite for video generative models.
\newblock In {\em Proceedings of the IEEE/CVF Conference on Computer Vision and Pattern Recognition}, 2024.

\bibitem{webvid}
Max Bain, Arsha Nagrani, G{\"u}l Varol, and Andrew Zisserman.
\newblock Frozen in time: A joint video and image encoder for end-to-end retrieval.
\newblock In {\em IEEE International Conference on Computer Vision}, 2021.

\bibitem{caron2021emerging}
Mathilde Caron, Hugo Touvron, Ishan Misra, Herv{\'e} J{\'e}gou, Julien Mairal, Piotr Bojanowski, and Armand Joulin.
\newblock Emerging properties in self-supervised vision transformers.
\newblock In {\em Proceedings of the IEEE/CVF international conference on computer vision}, pages 9650--9660, 2021.

\bibitem{radford2021learning}
Alec Radford, Jong~Wook Kim, Chris Hallacy, Aditya Ramesh, Gabriel Goh, Sandhini Agarwal, Girish Sastry, Amanda Askell, Pamela Mishkin, Jack Clark, et~al.
\newblock Learning transferable visual models from natural language supervision.
\newblock In {\em International conference on machine learning}, pages 8748--8763. PMLR, 2021.

\bibitem{li2023amt}
Zhen Li, Zuo-Liang Zhu, Ling-Hao Han, Qibin Hou, Chun-Le Guo, and Ming-Ming Cheng.
\newblock Amt: All-pairs multi-field transforms for efficient frame interpolation.
\newblock In {\em Proceedings of the IEEE/CVF Conference on Computer Vision and Pattern Recognition}, pages 9801--9810, 2023.

\bibitem{schuhmann2022laion}
Christoph Schuhmann, Romain Beaumont, Richard Vencu, Cade Gordon, Ross Wightman, Mehdi Cherti, Theo Coombes, Aarush Katta, Clayton Mullis, Mitchell Wortsman, et~al.
\newblock Laion-5b: An open large-scale dataset for training next generation image-text models.
\newblock {\em Advances in Neural Information Processing Systems}, 35:25278--25294, 2022.

\bibitem{teed2020raft}
Zachary Teed and Jia Deng.
\newblock Raft: Recurrent all-pairs field transforms for optical flow.
\newblock In {\em Computer Vision--ECCV 2020: 16th European Conference, Glasgow, UK, August 23--28, 2020, Proceedings, Part II 16}, pages 402--419. Springer, 2020.

\bibitem{ke2021musiq}
Junjie Ke, Qifei Wang, Yilin Wang, Peyman Milanfar, and Feng Yang.
\newblock Musiq: Multi-scale image quality transformer.
\newblock In {\em Proceedings of the IEEE/CVF international conference on computer vision}, pages 5148--5157, 2021.

\bibitem{wu2022grit}
Jialian Wu, Jianfeng Wang, Zhengyuan Yang, Zhe Gan, Zicheng Liu, Junsong Yuan, and Lijuan Wang.
\newblock Grit: A generative region-to-text transformer for object understanding.
\newblock {\em arXiv preprint arXiv:2212.00280}, 2022.

\bibitem{li2023unmasked}
Kunchang Li, Yali Wang, Yizhuo Li, Yi~Wang, Yinan He, Limin Wang, and Yu~Qiao.
\newblock Unmasked teacher: Towards training-efficient video foundation models.
\newblock In {\em Proceedings of the IEEE/CVF International Conference on Computer Vision}, pages 19948--19960, 2023.

\bibitem{huang2023t2i}
Kaiyi Huang, Kaiyue Sun, Enze Xie, Zhenguo Li, and Xihui Liu.
\newblock T2i-compbench: A comprehensive benchmark for open-world compositional text-to-image generation.
\newblock {\em Advances in Neural Information Processing Systems}, 36:78723--78747, 2023.

\bibitem{huang2023tag2text}
Xinyu Huang, Youcai Zhang, Jinyu Ma, Weiwei Tian, Rui Feng, Yuejie Zhang, Yaqian Li, Yandong Guo, and Lei Zhang.
\newblock Tag2text: Guiding vision-language model via image tagging.
\newblock In {\em The Twelfth International Conference on Learning Representations}, 2023.

\bibitem{wang2023internvid}
Yi~Wang, Yinan He, Yizhuo Li, Kunchang Li, Jiashuo Yu, Xin Ma, Xinhao Li, Guo Chen, Xinyuan Chen, Yaohui Wang, et~al.
\newblock Internvid: A large-scale video-text dataset for multimodal understanding and generation.
\newblock {\em arXiv preprint arXiv:2307.06942}, 2023.

\bibitem{wang2023lavie}
Yaohui Wang, Xinyuan Chen, Xin Ma, Shangchen Zhou, Ziqi Huang, Yi~Wang, Ceyuan Yang, Yinan He, Jiashuo Yu, Peiqing Yang, et~al.
\newblock Lavie: High-quality video generation with cascaded latent diffusion models.
\newblock {\em arXiv preprint arXiv:2309.15103}, 2023.

\bibitem{gen2}
Patrick Esser, Johnathan Chiu, Parmida Atighehchian, Jonathan Granskog, and Anastasis Germanidis.
\newblock Structure and content-guided video synthesis with diffusion models.
\newblock In {\em Proceedings of the IEEE/CVF International Conference on Computer Vision}, pages 7346--7356, 2023.

\end{thebibliography}
}

\end{document}